\pdfoutput=1

\documentclass[11pt]{article}

\usepackage[final]{acl}

\usepackage{times}
\usepackage{latexsym}

\usepackage[T1]{fontenc}

\usepackage[utf8]{inputenc}

\usepackage{microtype}

\usepackage{inconsolata}

\usepackage{graphicx}

\usepackage{tikz}
\usepackage[edges]{forest}
\usepackage{graphicx}
\usepackage{tabularx}

\usepackage{multirow}
\usepackage{booktabs}
\usepackage{float}
\usepackage{times}
\usepackage{latexsym}
\usepackage{url}
\usepackage{forest}
\usepackage{booktabs}
\usepackage{multirow}
\usepackage{tabularx}
\usepackage{graphicx}
\usepackage{adjustbox}
\usepackage{colortbl}
\usepackage{pifont}
\usepackage{enumitem}

\usetikzlibrary{trees,positioning,shapes,shadows,arrows.meta}

\usepackage[edges]{forest}
\definecolor{hidden-draw}{RGB}{205, 44, 36}
\definecolor{hidden-blue}{RGB}{194,232,247}
\definecolor{hidden-orange}{RGB}{243,202,120}
\definecolor{hidden-yellow}{RGB}{242,244,193}
\definecolor{tree-level-1}{RGB}{245,20,85}
\definecolor{tree-level-2}{RGB}{246,86,118}
\definecolor{tree-level-3}{RGB}{248,177,193}
\definecolor{tree-leaf}{RGB}{176,230,198}

\usepackage{xspace,mfirstuc,tabulary}

\title{From Generation to Judgment: Opportunities and Challenges of LLM-as-a-judge}

\author{
  Dawei Li\textsuperscript{\ding{171}}, 
  Bohan Jiang\textsuperscript{\ding{171}},  
  Liangjie Huang\textsuperscript{\ding{168}}, 
  Alimohammad Beigi\textsuperscript{\ding{171}}, 
  \textbf{Chengshuai Zhao}\textsuperscript{\ding{171}},\\
  \textbf{Zhen Tan}\textsuperscript{\ding{171}},  
  \textbf{Amrita Bhattacharjee}\textsuperscript{\ding{171}}, 
  \textbf{Yuxuan Jiang}\textsuperscript{\ding{169}}, 
  \textbf{Canyu Chen}\textsuperscript{\ding{166}}, 
  \textbf{Tianhao Wu}\textsuperscript{\ding{167}},\\ 
  \textbf{Kai Shu}\textsuperscript{\ding{165}}, 
  \textbf{Lu Cheng}\textsuperscript{\ding{168}}, 
  \textbf{Huan Liu}\textsuperscript{\ding{171}}\\
  \textsuperscript{\ding{171}}Arizona State University, 
  \textsuperscript{\ding{168}}University of Illinois Chicago,\\
  \textsuperscript{\ding{169}}University of Maryland, Baltimore County, 
  \textsuperscript{\ding{166}}Northwestern University,\\
  \textsuperscript{\ding{167}}University of California, Berkeley, 
  \textsuperscript{\ding{165}}Emory University
}

\begin{document}
\maketitle
\begin{abstract}
  Assessment and evaluation have long been critical challenges in artificial intelligence (AI) and natural language processing (NLP). Traditional methods, usually matching-based or small model-based, often fall short in open-ended and dynamic scenarios. Recent advancements in Large Language Models (LLMs) inspire the ``LLM-as-a-judge'' paradigm, where LLMs are leveraged to perform scoring, ranking, or selection for various machine learning evaluation scenarios. This paper presents a comprehensive survey of LLM-based judgment and assessment, offering an in-depth overview to review this evolving field. We first provide the definition from both input and output perspectives. Then we introduce a systematic taxonomy to explore LLM-as-a-judge along three dimensions: \textit{what} to judge, \textit{how} to judge, and \textit{how} to benchmark. Finally, we also highlight key challenges and promising future directions for this emerging area\footnote{More resources on \textbf{LLM-as-a-judge} are on the website: \url{https://llm-as-a-judge.github.io}}\footnote{We have released and will maintain a paper list about \textbf{LLM-as-a-judge} at: \url{https://github.com/llm-as-a-judge/Awesome-LLM-as-a-judge}}.
\end{abstract}

\section{Introduction}
\vspace{-1mm}

Automatic model assessment and evaluation have long been essential yet challenging tasks in machine learning (ML) and natural language processing (NLP)~\cite{sai2022survey,chang2024survey}.
Traditional static metrics~\cite{tan2022supervised,li2024quantmoe} like BLEU~\cite{papineni2002bleu} and ROUGE~\cite{lin2004rouge} measure quality by calculating lexical overlap between output and reference texts. While computationally efficient, these metrics perform poorly in dynamic and open-ended scenarios~\cite{liu2016not,reiter2018structured}.
With the rise of deep learning, small language model-based metrics like BERTScore~\cite{zhangbertscore} and BARTScore~\cite{yuan2021bartscore} have emerged. However, these metrics still face challenges in capturing nuanced attributes like fairness~\cite{sun2022bertscore} and helpfulness~\cite{zhu2024starling}.


Recently, the advancements of large language models (LLMs)~\cite{zhao2025chain} such as GPT-4~\cite{achiam2023gpt} and o1~\cite{jaech2024openai}, have led to striking improvements in various applications, leveraging substantial prior knowledge in vast training corpora. This progress has motivated researchers to propose the concept of ``LLM-as-a-judge''~\cite{zheng2023judging,wang2023chatgpt,liu2023g,chiang2023closer}, where LLMs are used to assess the candidate outputs by assigning scores, producing rankings, or selecting the best options, based on various input formats (e.g., point- and pair-wise), given context and instruction.
The strong capability of LLMs combined with well-designed assessment pipelines~\cite{li2023prd,bai2024benchmarking} leads to fine-grained and human-like judgment for various evaluation applications, addressing the previous limitations.


Beyond evaluation, LLMs-as-a-judge has been adopted across the lifecycle for next generations of LLM developments and applications.
LLMs-as-a-judge is often used as a scalable way to provide supervisions for key development steps like alignment~\cite{lee2023rlaif}, retrieval~\cite{li2024dalk}, and reasoning~\cite{liang2023encouraging}.
LLM-as-a-judge also empowers LLMs with a series of advanced capabilities such as self-evolution~\cite{sun2024salmon}, active retrieval~\cite{li2024dalk}, and decision-making~\cite{yang2023auto}, driving their elevations from generative models to intelligent agents~\cite{zhuge2024agent}.
However, as the field develops rapidly, challenges like bias and vulnerability~\cite{koo2023benchmarking,park2024offsetbias,fu2024gptscore,huang2024limitations} are emerging. Therefore, a systematic review of both techniques and limitations is crucial for facilitating this field.

This survey delves into the details of LLM-as-a-judge, aiming to provide a systematic overview of LLM-based judgment systems. We start by formally defining LLM-as-a-judge with its diverse input and output formats (Section~\ref{Preliminary}). Next, we propose an in-depth and comprehensive taxonomy to address the three key questions (Section~\ref{Attributes},~\ref{Methodology}~\ref{sec:benchmark_llm_as_judge}):
\begin{itemize}[leftmargin=*,itemsep=1pt]
\vspace{-1mm}
    \item \textbf{Attribute: What to judge?} We outline six subtle attributes that are uniquely assessed by LLM-as-a-judge, including helpfulness, safety \& security, reliability, relevance, logical, and overall quality.
    \vspace{-6mm}
    \item \textbf{Methodology: How to judge?} We explore ten tuning and prompting methods for LLM-as-a-judge, including manual labeling, synthetic feedback, supervised fine-tuning, preference learning, swapping operation, rule augmentation, multi-agent collaboration, demonstration, multi-turn interaction, and comparison acceleration.
    \vspace{-1mm}
    \item \textbf{Benchmark: How to evaluate LLM-as-a-judge?} We categorize existing benchmarks for LLM-as-a-judge into four types: for general performance, bias quantification, challenging tasks, and domain-specific performance.
\end{itemize}
Finally, we discuss challenges and potential future directions for LLM-as-a-judge in Section~\ref{Challenges and Future Works}.

\noindent \textbf{Differences from Existing Surveys.} Existing concurrent surveys investigate LLM for the evaluation of natural language generation (NLG)~\cite{gao2024llm,li2024leveraging,gu2024survey}.
However, LLM-as-a-judge has been applied across a broader range of scenarios beyond evaluation, as we discussed, necessitating a systematic survey to categorize and summarize its various applications. 

\section{Preliminary}
\label{Preliminary}
\begin{figure}[h]
\vspace{-1mm}
 \centering
  \includegraphics[width=0.45\textwidth]{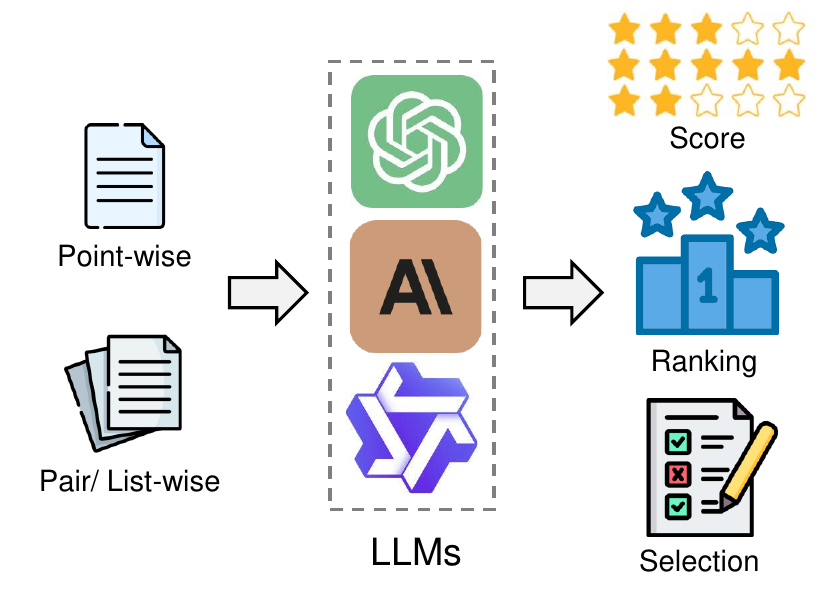}
  \caption{Overview of I/O formats of LLM-as-a-judge.}
  \label{fig:input-output}
  \vspace{-5mm}
\end{figure}

In this section, we provide a detailed definition of LLM-as-a-judge, discussing the various input and output formats as shown in Figure~\ref{fig:input-output}.

\subsection{Input}
\label{Input}
\vspace{-1mm}

Given a judge LLM $J$, the assessment process can be formulated as: $R = J(C_1,...C_n).$
Here $C_i$ is the $i_{th}$ candidate to be judged and $R$ is the judging result. We categorize two input formats based on the different candidate numbers $n$.

\noindent \textbf{Point-Wise:} When $n=1$, it becomes a point-wise judgment where the LLMs judges will solely focus on one candidate sample~\cite{gao2023human}.

\noindent \textbf{Pair/ List-Wise:} When $n\geq2$, it becomes a pair-wise ($n=2$) or list-wise ($n>2$) judgment where multiple candidate samples are provided together for the LLM judges to compare and make a comprehensive assessment~\cite{zheng2023judging}.

\subsection{Output}
\label{Output}

In this section, we discuss three kinds of output of the judgment based on the different formats of $R$.

\noindent \textbf{Score:} When each candidate sample is assigned a continuous or discrete score: $R=\{C_1:S_1,...,C_n:S_n\}$, it becomes a score-based judgment. This is the most widely adopted protocol, leveraging LLMs to generate scores for quantitative comparisons~\cite{li2024exploring} or attribute detection~\cite{xie2024sorry}.

\noindent \textbf{Ranking:} In ranking-based judgment, the output is a ranking of each candidate sample, represented as $R=\{C_i>...>C_j\}$. This comparative approach is useful in scenarios where establishing a rank order among candidates is required~\cite{li2023prd,liu2024aligning}.

\noindent \textbf{Selection:} In selection-based judgment, the output involves selecting one or more optimal candidates, represented as $R=\{C_i,...,C_j\} > \{C_1,...C_n\}$. This method is particularly crucial in decision-making~\cite{yao2023tree} or content-filtering~\cite{li2024dalk} contexts.

\section{Attribute}
\label{Attributes}
\begin{figure}[h]
 \centering
  \includegraphics[width=0.40\textwidth]{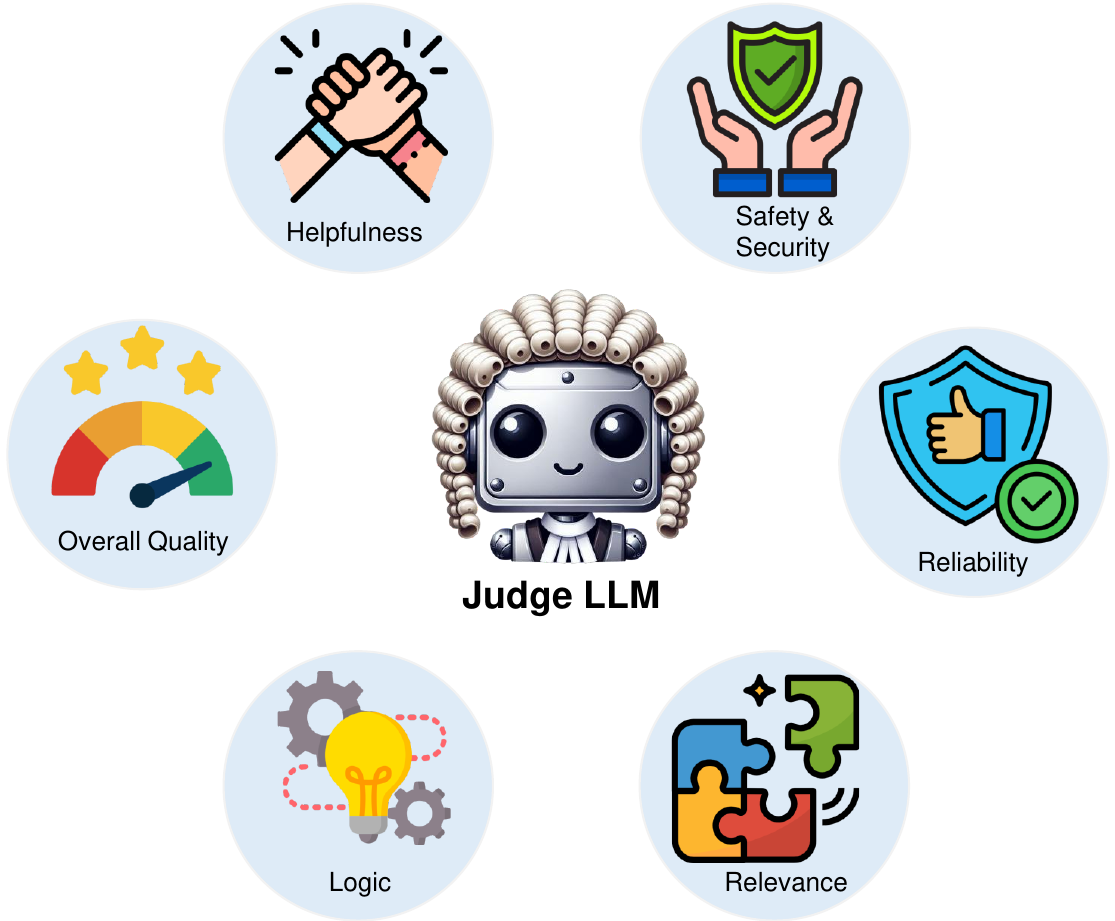}
  \caption{Overview of different judging aspects.}
  \label{fig:attribute}
\end{figure}

In this section, we categorize current research in LLM-as-a-judge from attribute perspectives. Figure~\ref{fig:attribute} gives an overview summarization of what aspects can be assessed by the LLM judges.

\subsection{Helpfulness}
\label{Helpfulness}


Helpfulness is a critical criterion to measure the utility and informativeness of a generated response.
Due to the high cost of manually assessing helpfulness in training data, recent studies have explored leveraging LLMs to label helpfulness and to generate or filter alignment data~\cite{bai2022constitutional,lee2023rlaif,guo2024direct,zhang2025persona}.
Beyond alignment tuning, helpfulness assessment using LLM-as-a-judge also plays a vital role in automatic model evaluation~\cite{zheng2023judging,lin2023unlocking,li2024generative,zhang2025sentient}.

\subsection{Safety \& Security}
\label{Harmlessness}

Safety and security are essential to ensure that models do not generate harmful content or respond inappropriately to malicious inputs.
Current studies have validated that LLMs can be effectively used for model safety assessment, either as off-the-shelf models guided by policy instructions~\cite{bai2022constitutional,phute2023llm,lirain,ye2024flask,wang2024not,eiras2025know,chen2025safer,rodriguez2025towards,hengle2025cseval}, or as lightweight models fine-tuned on safety-specific datasets~\cite{inan2023llama,zhang2024reviseval,xie2024sorry}.
Besides, LLM-as-a-judge has been widely adopted to detect and purify adversarial and toxic prompts designed with malicious intent~\cite{cantini2025benchmarking,mu2025evaluate,armstrong2025defense}.

\subsection{Reliability}
\label{Reliability}

Reliability is a crucial attribute for LLMs, enabling them to generate faithful content while presenting uncertainty or acknowledging missing knowledge about certain topics.
Regarding sentence-level faithfulness assessment, existing researches leverage LLM-as-a-judge to either instruct the powerful LLMs (e.g., GPT-4) directly~\cite{cheng2023evaluating,gekhman2023trueteacher,luo2024halludial,hsu2024rate} or train specific reliability judges~\cite{wang2024halu}.
Several works adopt LLM judges for long-form and fine-grained faithfulness evaluation~\cite{tan2024proxyqa,bai2024longbench,wu2025longeval}, using external retrieval bases~\cite{min2023factscore,cao2025video,loru2025decoding} or search engines~\cite{wei2024long}.
~\citet{jing2024faithscore,pu2025judge} further expand this assessment to the multimodal area.
Besides evaluation, there are also many works that adopt LLM-as-a-judge to improve the reliability of the generated content, either by external verifiers~\cite{xie2024improving} or synthetic alignment datasets~\cite{zhang2024self,wen2024policy}.
For uncertainty judgment,~\citet{xu2024sayself} propose SaySelf, a training framework that teaches LLMs to express more fine-grained confidence estimates with self-consistency prompting and group-based calibration training.

\subsection{Relevance}
\label{Relevance}

Relevance assessment with LLM-as-a-judge has been explored and validated to be a more refined and effective manner across various tasks~\cite{chiang2023can,arabzadeh2025benchmarking}.
In conversation evaluation, both~\citet{lin2023llm} and~\citet{abbasiantaeb2024can} propose to replace expensive human annotation with LLM judgment in relevance assessment.
In retrieval-augmented generation (RAG) scenarios, there are also many works that utilize LLMs to determine which demonstrations~\cite{li-qiu-2023-mot} or documents~\cite{li2024dalk} are most relevant for solving the current problem.
Recently, LLM-as-a-judge has also been used in multimodal applications for cross-modality relevance judgment~\cite{lee2024fleur,chen2024mj,yang2024toward,chen2024mllmasajudge,lu2024llmscore,luo2024videoautoarena,lin2025evaluating}.
Additionally, LLM-as-a-judge has also been explored in many traditional retrieval applications for relevance assessment~\cite{zhao2023llm,alaofi2024generative,dietz2025llm,arabzadeh2025human,balog2025rankers}, such as search~\cite{thomas2024large,sebastian2025validating}, retrieval~\cite{ma2024leveraging,dey2025judge}, recommendation~\cite{hou2024large,zhang2024largeas}.

\subsection{Logic}
\label{Feasibility}


In agentic LLMs, assessing the logical correctness of candidate actions or steps is crucial for LLMs' planning, reasoning and decision-making, which further releases their great potential at inference-time.
While some works leverage metrics or external tools for this feasibility assessment~\cite{huang2023large,yuan2024advancing}, many others leverage LLMs' feedback as the signal~\cite{lightmanlet,kawabata2024rationale} to perform planning and searching in complex reasoning spaces~\cite{hao2023reasoning,yao2023tree,besta2024graph}.
In multi-agent collaboration systems, both~\citet{liang2023encouraging} and~\citet{li2024smoa} propose to leverage the judge LLM to select the most feasible solutions among multiple candidates' responses.
Besides, other works adopt LLM judges to perform logical assessment in API selection~\cite{zhao2024diffagent}, tool using~\cite{yang2023auto} and LLM routing~\cite{ong2024routellm}.

\subsection{Overall Quality}
\label{Overall Quality}


As previously mentioned, LLM-as-a-judge can be employed to perform multi-aspect and fine-grained assessments. However, in many cases, a general assessment is still required to represent the candidates' overall quality.
One straightforward approach to obtain this overall score is based on the aspect-specific scores, either by averaging them~\cite{lin2023unlocking} or referring them to generate an overall judgment~\cite{yu2024kieval}.
Moreover, in many traditional NLP tasks~\cite{lu2024beyond,jiang2024genres,ho2025llm,shibata2025lces,kartavc2025openlgauge} like summarization~\cite{gao2023human,Jain2023MultiDimensionalEO,chen2023exploring,kumar2024llms,qi2025evaluating,barnes2025summarization,altemeyer2025argument,jeong2025agent,calderon2025alternative} and machine translation~\cite{Kocmi2023LargeLM,Huang2024LostIT,piergentili2025llm,wang2025contrastscore}, the evaluation dimensions are less diverse compared to more open-ended, long-form generation tasks. As a result, LLM-as-a-judge is often prompted directly to produce an overall judgment in these tasks.

\section{Methodology}
\label{Methodology}

In this section, we present commonly adopted methods and tricks to improve LLMs' judging capabilities, splitting them into tuning (Section~\ref{Tuning}) and prompting strategies (Section~\ref{Prompting}).

\subsection{Tuning}
\label{Tuning}
To enhance the judging capabilities of a general LLM, various tuning techniques have been employed in different studies. In this section, we discuss these tuning approaches for LLM-as-a-judge from two perspectives: data sources (Section~\ref{Data Source}) and training techniques (Section~\ref{Tuning Techniques}). 

\subsubsection{Data Source}
\label{Data Source}

\paragraph{Manually-labeled Data:}\label{manually} 

To train a LLM judge with human-like criteria, one intuitive method is to collect manually-labeled judgments.
Previous works have leveraged and integrated existing sources annotated by humans, including instruction tuning datasets~\cite{lee2024aligning,wang2024pandalm} and traditional NLP datasets~\cite{vu2024foundational}, for tuning LLM judges.
Other works collect manually-labeled datasets with fine-grained judgment feedback.
These fine-grained feedbacks can be rationales behind judgment results~\cite{xu2023instructscore}, multi-aspect judgment formats~\cite{liu2024x} and fine-grained judgment labels~\cite{yue2023automatic}, all of which facilitate the LLM judges to produce more detailed and context-rich judging results. 
Notably,~\citet{ke2024critiquellm} first prompt GPT-4 to generate judgment and then manually verify and revise the outputs to ensure high-quality annotations.

\paragraph{Synthetic Feedback:}\label{synthetic}

While manually labeled feedback is high-quality and accurately reflects human judgment preferences, it is limited in both scale and coverage. To address it, researchers have also explored synthetic feedback as a data source for LLM judges' tuning.
Some rely on the LLM judges themselves to generate the synthetic feedback.
It involves instructing the LLM to self-evaluate and improve its judgments~\cite{wu2024meta}, or by generating corrupted instructions and corresponding responses as negative samples for Directed Preference Optimization (DPO) training~\cite{wang2024self}.
Besides, other powerful and stronger LLMs are also introduced for feedback synthesis.
For example, GPT-4 has been widely leveraged to synthesize judging evidence~\cite{wang2024halu}, erroneous responses~\cite{park2024offsetbias}, rationale and feedback~\cite{li2024generative,kim2024prometheus,xiong2024llava}, and judgment labels~\cite{zhu2023judgelm,xie2024sorry}.

\subsubsection{Tuning Techniques}
\label{Tuning Techniques}

\paragraph{Supervised Fine-tuning:}\label{sft}

Supervised fine-tuning (SFT) is the most widely used approach for training LLM judges~\cite{hu2025training}, enabling them to learn from pairwise~\cite{li2024generative,wang2023learning,zhu2023judgelm,wang2024pandalm,pombal2025m,salinas2025tuning} or pointwise~\cite{wang2023learning,yue2023automatic,xie2024sorry,lee2024aligning,chiang2025tract} judgment data.
Among many tricks applied in SFT, multi-task training and weight merging are introduced to enhance the robustness and generalization of LLM judges~\cite{kim2024prometheus,vu2024foundational,saad2024lmunit}.
Other works try to enrich the original training set with augmented or self-generated samples.
\citet{ke2024critiquellm} augment pairwise training data by swapping the order of two generated texts and exchanging the corresponding content in critiques.
\citet{xu2023instructscore} further fine-tune their INSTRUCTSCORE model on self-generated outputs to align diagnostic reports better with human judgment.
Additionally, \citet{liu2024x} propose a two-stage SFT process: an initial phase of vanilla instruction tuning for evaluation diversity, followed by additional tuning with auxiliary aspects to enrich the model's evaluative depth.

\begin{figure*}[h]
    \centering
    \includegraphics[width=0.8\linewidth]{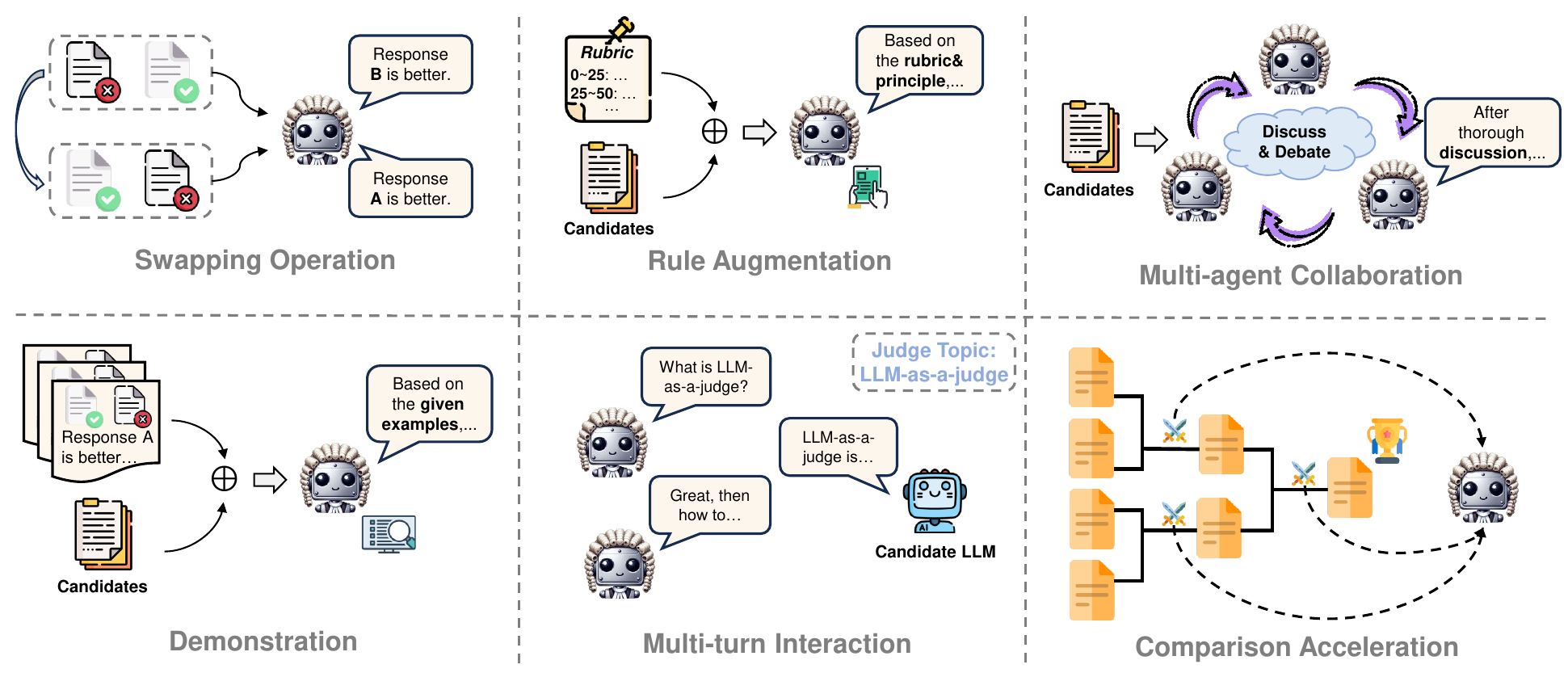}
    \caption{Overview of prompting strategies for LLM-as-a-judge.}
    \label{fig:prompting}
\end{figure*}

\paragraph{Reinforcement Learning:}\label{pl} 


Reinforcement learning from human preference is closely tied to judgment and evaluation tasks, particularly those involving comparison and ranking. Rather than directly adopt or augment preference learning datasets for SFT, several studies apply preference learning techniques to enhance LLMs' judging capabilities.
One straightforward way is to treat the off-topic responses as inferior samples and apply DPO~\cite{wang2024halu,yu2025improve,rad2025refining}.
Besides,~\citet{wu2024meta} propose meta-rewarding, which leverages the policy LLMs to judge the quality of their own judgment and produce pairwise signals for enhancing the LLMs' judging capability.
This concept is also adopted by \citet{wang2024self}, who propose self-taught evaluators that use corrupted instructions to generate suboptimal responses as inferior examples for preference learning.
Moreover,~\citet{hu2024themis} introduce rating-guided DPO, in which the rating difference between two responses is considered in preferences modeling.
Different from RLHF- and DPO-based approaches, several recent works leverage reinforcement learning with verifiable reward (RLVR)~\cite{guo2025deepseek} to train LLM judges by rewarding reasoning trajectories that lead to correct judgments~\cite{saha2025learning,liu2025inference,zhou2025evaluating}.

\subsection{Prompting}

Designing appropriate prompting strategies and pipelines at the inference stage could improve judgment accuracy and mitigate bias. We summarize existing prompting strategies for LLM-as-a-judge into six categories (see Figure~\ref{fig:prompting}).

\label{Prompting}
\subsubsection{Swapping Operation}\label{swapping}

Previous studies have demonstrated that LLM-based judges are sensitive to the positions of candidates, and the ranking results of candidate responses can be easily manipulated by merely altering their order in the context~\cite{wang2023large}. To mitigate this positional bias and establish a more fair LLM judging system,~\cite{zheng2023judging} propose a swapping operation, which involves invoking the judge LLM twice, swapping the order of the two candidates in each instance. If the two results are inconsistent, it is labeled a ``tie'', indicating that the LLM is unable to confidently distinguish the quality of the candidates.
This swapping operation technique has also been widely adopted in pairwise feedback synthesis to produce more accurate reward signals~\cite{lee2023rlaif,sun2024salmon,lee2024aligning}.

\subsubsection{Rule Augmentation}\label{rule}

Rule-augmented prompting involves embedding a set of principles, references, and evaluation rubrics directly within the prompt for LLM judges. 
This approach is commonly employed in LLM-based evaluations, where LLM judges are guided to assess specific aspects~\cite{lahoti2023improving,li2024generative,bai2024benchmarking,yu2024kieval,qian2024large,dong2024can,wei2025rocketeval,xie2025prompting} and provided with detailed rubrics and criteria~\cite{gao2023human,kim2023prometheus,wang2024ceb,murugadoss2024evaluating,li2024rule,li2024decompose,hu2024llm,liu2024hd,li2025hypoeval,fan2025sedareval} to ensure accurate judgments.
Following this concept, studies in alignment~\cite{bai2022constitutional,lee2023rlaif,lee2024aligning,guo2024direct,sun2024salmon,beigi2024lrq} enhance this principle-driven prompting by incorporating more detailed explanations~\cite{hueditable} for each aspect of the principle or rubric.
Apart from these human-written rules, some works~\cite{liu2024calibrating,zhang2024reviseval,xu2025learning,wen2025hpss,zhou2024fairer} embed the self-generated or automaticaly-searched scoring criteria and principles as a part of their instruction.

\subsubsection{Multi-agent Collaboration}\label{multi}

Accessing results from a single LLM judge may not be reliable due to inherent biases in LLMs~\cite{wang2023large, liusie2024llm, ohi2024likelihood}. To address this limitation, \citet{li2023prd,chen2024automatic,ning2024pico} introduce the Peer Rank (PR) algorithm, which produces the final ranking based on each LLM judge's output. Building on this, several architectures and techniques for multi-agent LLMs emerge, including mixture-of-agent~\cite{zhang2023wider,xu2023towards,beigi2024lrq,cao2025multi}, role play~\cite{wu2023large,li2024mateval,patel2024aime}, debating~\cite{chanchateval,Zhang2024BreakingER,bandi2024adversarial,kenton2024scalable}, voting \& aggregation~\cite{zhu2024dynamic,verga2024replacing,li2025leveraging,guerdan2025validating,rahmani2024judgeblender} and cascaded selection~\citet{jung2024trust,badshah2025dafe}. Additionally, others apply multi-agent collaboration for alignment data synthesis, leveraging multiple LLM judges to refine responses~\cite{arif2024fellowship} or provide more accurate feedback~\cite{li2024coevol}.

\subsubsection{Demonstration}\label{demon}


In-context samples or demonstrations~\cite{brown2020language, dong2022survey, agarwal2024many} provide concrete examples for LLMs to follow and have been shown to be a crucial factor in the success of in-context learning for LLMs. Several studies have introduced human assessment results as demonstrations for LLMs-as-judges, aiming to help LLMs learn evaluation standards from a few illustrative examples~\cite{jain2023multi,kotonya2023little}.
To improve the robustness of LLM evaluations, \citet{hasanbeig2023allure} propose ALLURE, an approach that iteratively incorporates demonstrations of significant deviations to enhance the evaluator’s robustness. Additionally, \citet{song2024can} borrow the insights from many-shot in-context learning and apply it in LLM-as-a-judge applications.

\subsubsection{Multi-turn Interaction}\label{multiturn}


A single response may not provide enough information for an LLM judge to thoroughly and fairly assess each candidate. To address this limitation, multi-turn interactions are proposed to offer a more comprehensive evaluation. 
Typically, the process begins with an initial query or topic, followed by dynamically interacting between the LLM judge and candidate models~\cite{bai2023benchmarking,yu2024kieval,pombal2025zero}.
Besides, some approaches facilitate debates among candidates in a multi-round manner, allowing their true knowledge and performance to be fully revealed and evaluated~\cite{zhao2024auto,moniri2024evaluating}.

\subsubsection{Comparison Acceleration}\label{compare}


Among various input formats in LLM-as-a-judge, pair-wise comparison is the most common approach for model comparison in evaluation or producing pair-wise feedback for training. However, when multiple candidates need to be ranked, this method can be quite time-consuming~\cite{zhai2024online}. To mitigate the computational overhead, \citet{zhai2024online} propose a ranked pairing method in which all candidates are compared against an intermediate baseline response. In addition, \citet{lee2024aligning,liu2025pairwise} utilize a tournament-based approach~\cite{liu2023statistical, zhao2023slic} for rejection sampling during inference to speed up the pair-wise comparison.

\section{Application}
\label{Application}
We introduce four applications which LLM-as-a-judge can be applied: evaluation (Section~\ref{Evaluation}), alignment (Section~\ref{Alignment}), retrieval (Section~\ref{Retrieval}), and reasoning (Section~\ref{Reasoning}). Due to the space limitation, we provide a more detailed version in Appendix~\ref{Application with More Details}.

\subsection{Evaluation}
\label{Evaluation}
LLM judges are initially proposed for and widely adopted in various evaluation scenarios. For open-ended generation, LLM judges assess the quality of outputs like dialogues, summaries, and creative writing, ensuring contextual relevance, coherence, and safety~\cite{badshah2024reference,kumar2024decoding,zengevaluating,jones2024multi}. For reasoning tasks, they judge intermediate steps and final answers~\cite{he2023socreval,parmar2024logicbench,xia2024evaluating} in areas such as math~\cite{xia2024evaluating}, logic~\cite{parmar2024logicbench}, and temporal reasoning~\cite{fatemi2024test}. There are also some emerging areas where LLM judges are applied to domains once dominated by humans, including social intelligence~\cite{zhou2023sotopia}, multimodal tasks~\cite{chen2024mllm} and multilingual generation~\cite{fu2025reliable}.

\subsection{Alignment}
\label{Alignment}
Model alignment also benefits from the automatic LLM-as-a-judge to produce and filter data at scale. Typically, larger and powerful LLMs are usually used as judges to align smaller models, providing synthetic preference data. This includes methods like multi-agent collaboration~\cite{arif2024fellowship} and specialized tasks such as code alignment~\cite{weyssow2024codeultrafeedback}. Additionally, self-judging methods have LLMs rank or critique their own outputs to generate preference data without external teachers. To improve the judging capability of the policy model, techniques such as meta-rewarding~\cite{wu2024meta}, Judge Augmented Supervised Fine-Tuning (JSFT)~\cite{lee2024aligning}, and self-evaluation~\cite{zhang2024self} have been proposed. Apart from pairwise data, some other studies also use LLM-as-a-judge to judge and filter synthetic SFT data for instruction tuning~\cite{liang2024sheep,yasunaga2024alma}.

\subsection{Retrieval}
\label{Retrieval}
LLM judges can assist with both traditional retrieval tasks and retrieval-augmented generation (RAG). For traditional retrieval, LLM-as-a-judge ranks documents by relevance~\cite{zhuang2024beyond} without task-specific data~\cite{ma2023zero}, using permutation-based~\cite{sun2023chatgpt}, pairwise~\cite{qin2024large}, and listwise~\cite{zhuang2024setwise} approaches to improve reranking for complex queries and domain-specific search tasks. For RAG, LLM judges guide how external knowledge is fetched and used during generation, ensuring coherence, accuracy, and relevance. This includes frameworks like Memory-of-Thought~\cite{li2023mot}, Self-Retrieval~\cite{tang2024self}, and Self-RAG~\cite{asaiself}, where the judge selects or filters retrieved content, particularly in specialized fields such as biomedicine~\cite{li2024dalk}.

\subsection{Reasoning}
\label{Reasoning}
Reasoning is a critical capability of LLMs for complex and dynamic problem-solving. LLM judges can aid reasoning tasks by improving reasoning path selection and external tool use. Reasoning path selection involves identifying the correct trajectory for the LLM’s reasoning process, where LLM-as-a-judge are adopted to evaluate intermediate reasoning steps~\cite{lahoti2023improving}, perform trajectory-level selection~\cite{musolesi2024creative}, and act as a process reward model for reasoning state scoring~\cite{lightman2023let} or a fine-grained critic to provide verbal feedback~\cite{ankner2024critique}. For external tool use, LLM judges help AI systems decide which external tools, modules, or agents to activate at each step of reasoning, acting as controllers that coordinate tool choice~\cite{sha2023languagempc}, agent communication~\cite{ong2024routellm}, and message flow management~\cite{liang2023encouraging} to ensure accurate and coherent problem solving.

\section{Benchmark: Judging LLM-as-a-judge}
\label{sec:benchmark_llm_as_judge}



We categorize benchmarks for evaluating LLMs-as-judges into four groups: general performance (Section~\ref{General Performance}), bias quantification (Section~\ref{Bias Quantification}), challenging task performance (Section~\ref{Challenging Task Performance}), and domain-specific performance (Section~\ref{Domain-Specific Performance}).

\subsection{General Performance}
\label{General Performance}

Benchmarks focusing on general performance aim to evaluate the overall competence of LLMs in various tasks.
One direct way to benchmark LLM judges' performance is to calculate the alignment between LLM prediction and the manual judgment result, using various metrics like Cohen’s kappa, Discernment Score, and normalized accuracy~\cite{li2023exploring,tan2024judgebench,wang2024dhp,lambert2024rewardbench,penfever2024sosbench,qu2025efficient,xu2025does,chang2025exploring,hu2025dual,calderon2025alternative,elangovan2024beyond,schroeder2024can,gera2024justrank}.
Moreover, several studies build LLM leaderboards using LLM-as-a-judge and assess their validity by comparing model rankings with those from established benchmarks and leaderboards, such as Chatbot Arena~\cite{zheng2023judging})~\cite{zheng2023judging,dubois2024length,li2024crowdsourced,zhao2024auto,chi2025copilot}.


\subsection{Bias Quantification}
\label{Bias Quantification}
Quantifying and mitigating bias in LLM judgments is critical to ensuring fairness and reliability~\cite{xie2025empirical}. Typical benchmarks include EvalBiasBench~\cite{park2024offsetbias} and CALM~\cite{ye2024justice}, focus explicitly on quantifying biases, including those emerging from alignment and robustness under adversarial conditions. Besides, \citet{shi2024optimization} adopt metrics such as position bias and percent agreement in question-answering tasks.
Recently,~\cite{tripathi2025pairwise} examine the influence of protocol choice (pairwise and pointwise) on the bias degree of LLM judges.

\subsection{Challenging Task Performance}
\label{Challenging Task Performance}
Benchmarks designed for difficult tasks push the boundaries of LLM evaluation. For example, Arena-Hard \cite{li2024crowdsourced} and JudgeBench \cite{tan2024judgebench} select harder questions based on LLMs' performance for conversational QA and various reasoning tasks, respectively. CALM \cite{ye2024justice} explores alignment and challenging scenarios, using metrics like separability and agreement to evaluate performance in manually identified hard datasets.

\subsection{Domain-Specific Performance}
\label{Domain-Specific Performance}
Domain-specific benchmarks provide task-focused evaluations to assess LLMs’ effectiveness in specialized contexts. Concretely, \citet{raju2024constructing} measure separability and agreement across tasks in specific domains such as coding, medical, finance, law and mathematics. CodeJudge-Eval \cite{zhao2024codejudge} specifically evaluates LLMs for judging code generation with execution-focused metrics such as accuracy and F1 score.
This idea has also been adopted by several following works in code summarization and generation evaluation~\cite{Wu2024CanLL,Yang2024EvaluatingAA,tong2024codejudge}.
Besides, there are also domain-specific benchmarks focusing on LLMs' assessing capabilities in multimodal~\cite{chen2024mllmasajudge}, multilingual~\cite{son2024mmeval,son2024llm}, instruction following~\cite{murugadoss2024evaluating} and LLM agent~\cite{lu2025agentrewardbench}.

\section{Challenges \& Future Works}
\label{Challenges and Future Works}

\subsection{Bias \& Vulnerability}
\label{bias}

The use of LLMs-as-a-judge inherently introduces significant challenges related to bias and vulnerability, which significantly compromise fairness and reliability when LLMs are deployed for diverse judging tasks.
Among the various types of bias, some are consistent across all LLM judges, for example, a tendency to prefer longer~\cite{koo2023benchmarking,dubois2024length,domhan2025same,yuan2023batcheval}, authoritative-looking~\cite{stephan2024calculation,chen2024humans} and well-formatted~\cite{chen2024humans} responses.
In addition, other biases stem from individual judges’ own preferences or knowledge, such as egocentric bias~\cite{liu2023llms, wataoka2024self,panickssery2024llm,chen2025llm} and preference leakage~\cite{li2025preference,goel2025great,naseh2025synthetic}.
LLM judges are also susceptible to adversarial manipulations. Techniques like prompt injection attacks~\cite{shi2024optimization,benchmarkjailjudge,Banerjee2024TheVO,tong2025badjudge} and adversarial phrases~\cite{liusie2023zero, raina2024llm, doddapaneni2024finding} can drastically influence LLMs' judgment, thus raising concerns about the reliability of LLM judges in high-stakes scenarios~\cite{shi2024optimization, raina2024llm}.

\noindent\textbf{Future Direction.} Existing studies have already explored approaches, such as providing more detailed evaluation principles~\cite{zheng2023judging,zhu2023judgelm,liusie2023zero,krumdick2025no} and eliminating spurious features through calibration~\cite{li2024calibraeval,raina2024llm,zhou2024mitigating,liu2024calibrating,chen2025unbiased,wang2025improving,van2025aligning}, to mitigate LLM judges' bias. Future work could focus more on analyzing and understanding the \textbf{root causes} of these biases. For example, why would LLMs prefer their own generation~\cite{panickssery2024llm}?

\subsection{Scaling Judgment at Inference Time.} Motivated by recent inference-time scaling (ITS) studies in LLMs~\cite{snell2024scaling,zhang2025and}, several works have begun to explore how to scale LLMs' judgment capabilities at inference time~\cite{saha2025learning,liu2025inference,zhou2025evaluating}. By expanding the reasoning process in judgment tasks and incorporating advanced behaviors such as reflection and exploration, both the accuracy and fairness~\cite{chen2025llm,wang2025assessing} of judge LLMs have seen significant improvements. A straightforward approach to scaling judge LLMs is to employ Large Reasoning Models (LRMs) that generate judgments via long CoT reasoning~\cite{chen2025judgelrm}. Additionally, traditional sampling and search strategies, such as self-consistency, best-of-N, and Monte Carlo Tree Search (MCTS), have been used to more thoroughly explore the space of possible judgment trajectories~\cite{wang2025mcts,kalra2025verdict}. Other methods leverage golden labels as supervision, applying rule-based reinforcement learning~\cite{chen2025judgelrm,liu2025inference,whitehouse2025j1incentivizingthinkingllmasajudge,chen2025rm,shi2025heimdall}, DPO~\cite{saha2025learning} or distillation~\cite{zhao2025genprm} to train LLMs to serve as more effective judges.

\noindent\textbf{Future Directions.} While LLM-as-a-judge approaches benefit from ITS techniques, it is also important to recognize the associated challenges. These include efficiency bottlenecks~\cite{sui2025stop}, performance degradation from over-thinking~\cite{chen2024not}, and increased vulnerability of long CoTs to adversarial attack~\cite{jiang2025safechain}. Future research could investigate these limitations and develop mitigation strategies, paving the way for more efficient, accurate, and robust judge LLMs enhanced by ITS.

\subsection{Dynamic \& Complex Judging Strategy}

Compared with earlier static and straightforward approaches that directly prompt LLMs for judgment~\cite{zheng2023judging}, more dynamic and complex judgment pipelines have been proposed recently to address various limitations, improving the robustness and effectiveness of LLM-as-a-judge.
One approach is to follow the concept of ``\textbf{LLM-as-a-examiner}'', where the system dynamically and interactively generates both questions and judgments based on the candidate LLMs' performance~\cite{yu2024kieval,bai2024benchmarking,pombal2025zero,dammu2025dynamic,khalili2025autodrive,wang2024revisiting,kim2024llm,zhang2025can}.
Other works focus on making judgments based on multiple candidate LLMs' \textbf{battling and debating}~\cite{moniri2024evaluating,zhao2024auto}.
Additionally, building complex judgment agents is another popular research area~\cite{li2023prd,chanchateval,zhuge2024agent}, which typically involves multi-agent collaboration with well-designed planning systems.

\noindent\textbf{Future Direction.} One promising direction for future research is to equip LLMs with human-like and agentic judgment capabilities~\cite{yuan2023batcheval,liang2024abseval,li2024split,saha2024branch,zhang2024evaluation,wang2025codevisionary,song2025grp}, such as anchoring, comparing, and meta-judgment. Another intriguing avenue would be to develop an \textbf{adaptive difficulty assessment system}~\cite{hu2024developing,patel2025get}, dynamically adjusting problems' difficulties based on candidates' performance.

\subsection{Human-LLMs Co-judgement}



As mentioned earlier, the biases and vulnerabilities in LLM-as-a-judge can be addressed through human-in-the-loop for further intervention and proofreading. However, only a few studies have focused on this direction~\cite{wang2023large,faggioli2023perspectives,pradeep2025great}.

\noindent\textbf{Future Direction.} As \textbf{data selection}~\cite{xie2023data, albalak2024survey} becomes an increasingly popular research area for improving the efficiency of LLMs' training and inference, it also holds the potential for enhancing LLMs-based evaluation. LLM-as-a-judge can draw insights from data selection to enable judge LLMs to serve as a critical sample selector, choosing a small subset of samples based on specific criteria (e.g., difficulty) for human annotators to conduct evaluation.

Due to the space limitation, we put the application of LLM-as-a-judge, paper collection for our taxonomy, tuning techniques and benchmark for LLM-as-a-judge in Appendix~\ref{Application},~\ref{Taxonomy},~\ref{Tuning Methods} and~\ref{Benchmark}.

\section{Conclusion}

This survey explores the intricacies of LLM-as-a-judge. We begin by categorizing existing LLM-based judgment methods based on input and output formats. Then, we propose a comprehensive taxonomy for LLM-as-a-judge, encompassing judging attributes, methodologies and benchmarks. After this, a detailed and thoughtful analysis of current challenges and future directions of LLM-as-a-judge is proposed, aiming to provide more resources and insights for future works in this emerging area.

\section*{Limitations}
This work aims to provide a comprehensive survey of the LLM-as-a-judge paradigm. Due to space constraints, we focus on three core aspects in the main paper: judging attributes, methods, and benchmarks. Applications of LLM-as-a-judge and a detailed list of related papers are included in the appendix. Additionally, as discussed in Section~\ref{bias}, LLM-as-a-judge carries inherent limitations and biases. The substantial computational resources required for deploying LLMs may also pose challenges in resource-constrained scenarios.

\section*{Acknowledgment}
This material is based upon work supported by the U.S. Department of Homeland Security under Grant Award Number 17STQAC00001-08-00. The views and conclusions contained in this document are those of the authors and should not be interpreted as necessarily representing the official policies, either expressed or implied, of the U.S. Department of Homeland Security. Lu Cheng is supported by the National Science Foundation (NSF) Grant \#2312862, NSF CAREER \#2440542, NSF-Simons SkAI Institute, National Institutes of Health (NIH) \#R01AG091762, Google Research Scholar Award, and a Cisco gift grant.

\bibliography{custom}

\appendix

\newpage

\section{Attribute Definition}
We provide a detailed definition for each judgment attribute in Table~\ref{tab:attributes_def}.

\begin{table*}[h!]
\centering
\small
\begin{tabular}{p{4cm} p{10cm}}
\toprule[1.2pt]
\textbf{Attribute} & \textbf{Definition} \\
\midrule
Helpfulness & Helpfulness is a critical criterion to measure the utility and informativeness of a generated response. \\
\addlinespace[4pt]
Safety \& Security & Safety \& security refer to whether the model avoids generating and is not affected by harmful, toxic, biased, or adversarial content. \\
\addlinespace[4pt]
Reliability & Reliability is the degree to which a response is faithful to verifiable sources and appropriately calibrated in expressing uncertainty. \\
\addlinespace[4pt]
Relevance & Relevance is a metric to measure how well a response aligns with the user query, topic, or task context. \\
\addlinespace[4pt]
Logic & Logic refers to the internal coherence and correctness of reasoning steps within a response, independent of factual accuracy. \\
\addlinespace[4pt]
Overall Quality & Overall quality is a holistic assessment of a response’s merit, typically integrating multiple dimensions into one comprehensive score. \\
\bottomrule[1.2pt]
\end{tabular}
\caption{Common judgment attributes and their definitions.}
\label{tab:attributes_def}
\end{table*}

\section{Prompting Methods Categories}
Based on each prompting strategy's target, we categorize them into following four group: (1) bias reduction, which involves reducing bias caused by candidate output position or reliance on a single LLM judge (swapping operations, multi-agent collaboration); (2) boosting instruction-following, which helps the LLM judge learn clear judging criteria and principles from rules or demonstrations (rule augmentation, in-context demonstration); (3) enhancing evaluation depth, which enables a better understanding of model capabilities (multi-turn interaction); and (4) improving evaluation efficiency, which refers to reducing the computational budget required during judgment (comparison acceleration).

\section{Application with More Details}
\label{Application with More Details}


\begin{figure*}[h]
    \centering
    \includegraphics[width=0.8\linewidth]{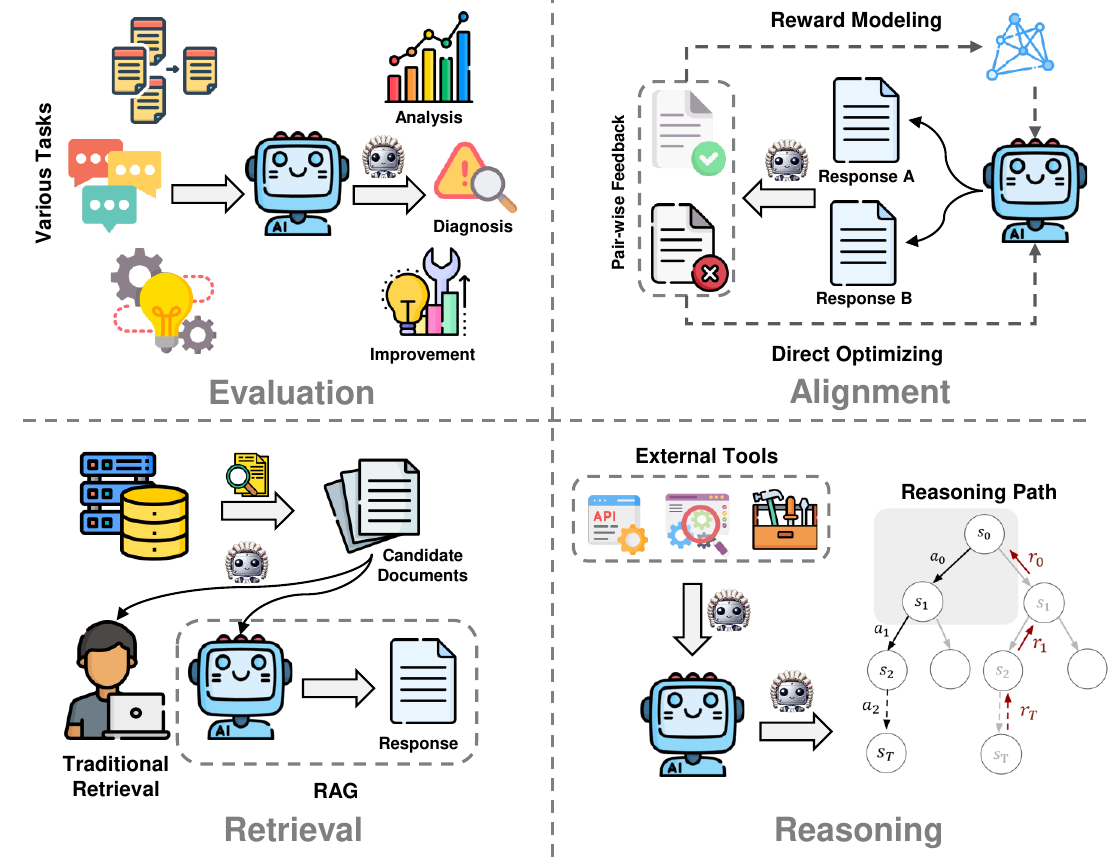}
    \caption{Overview of application and scenario for LLM-as-a-judge.}
    \label{fig:application}
\end{figure*}

\subsection{Evaluation}
LLM-as-a-judge is first proposed for evaluation. It enables human-like evaluations rather than overlap-based matching~\cite{post2018call, Lin2023LLMEvalUM}. We discuss how LLM-as-a-judge has been utilized to evaluate open-ended generation (Section~\ref{Open-ended Generation Tasks}), reasoning (Section~\ref{Reasoning Tasks}), and emerging NLP tasks (Section~\ref{Emerging Tasks}).

\subsubsection{Open-ended Generation Tasks}
\label{Open-ended Generation Tasks}

Open-ended generation includes tasks like dialog response, text summarization, and creative writing, where outputs must be safe, accurate, and contextually relevant with multiple ``correct'' answers~\cite{badshah2024reference, kumar2024decoding, zengevaluating,song2024finesure,jones2024multi}. Unlike traditional metrics, LLM-as-a-judge enables nuanced and adaptable evaluation~\cite{zheng2023judging}. This approach has been used for single-model evaluations and competitive comparisons~\cite{gao2023human, wu2023large}. While LLMs-as-judges demonstrate human-like judgments, longer outputs risk hallucinations~\cite{wang2024halu, cheng2023evaluating}. Another concern is biased and unsafe judgements~\cite{yu2024xfinder, li2024salad, ye2024justice}, though excessive caution may cause overly refusal~\cite{xie2024sorry}. To address these, researchers have proposed conversational frameworks like self-reflection~\cite{ji2023towards} and debating~\cite{moniri2024evaluating}. 
Besides, multilingual LLM-as-a-judge research has advanced with various methods and benchmarks that address cross-lingual evaluation challenges. Approaches include scoring non-English answers against English references~\cite{doddapaneni2024cross}, using multi-agent debate frameworks for fine-grained evaluation~\cite{feng2024m}, and developing open-source multilingual judges that outperform English-centric evaluators across 20+ languages~\cite{pombal2025m}. Benchmarks like MM-Eval and PARIKSHA test the consistency and fairness of multilingual LLM judges, showing that evaluators tuned in English often underperform on low-resource languages~\cite{son2024mmeval}.

However, key challenges still remain in LLM-based multilingual judgment. Studies highlight cross-lingual inconsistency, where judges show low agreement across languages, especially for low-resource settings~\cite{fu2025reliable}. Evaluators may also suffer from factual errors, cultural misrepresentations, and toxic content~\cite{hada2024large}. Additionally, dialectal variation further complicates the bias, with weaker alignment between LLM and human toxicity ratings in regional varieties [8]. These issues underscore the need for more culturally sensitive and robust multilingual evaluation methods.

\subsubsection{Reasoning Tasks}
\label{Reasoning Tasks}

The reasoning abilities of LLMs can be assessed through their intermediate thinking processes and final answers~\cite{he2023socreval,parmar2024logicbench, mondorf2024beyond}. For mathematical reasoning,~\citet{xia2024evaluating} introduce a framework using judge LLMs to assess the quality of reasoning steps. Similarly, for temporal reasoning,~\citet{fatemi2024test} create synthetic datasets to evaluate models' ability to reason about event sequences, causality, and dependencies. To distinguish genuine reasoning ability from pattern memorization,~\citet{wang2023can} propose a human-in-the-loop framework where LLMs and users adopt opposing positions to reach correct decisions.~\citet{nan2023evaluating} develop a multi-agent framework simulating peer review, leveraging LLMs-as-judges to collaboratively assess reasoning capabilities in data-driven tasks.

\subsubsection{Emerging Tasks}
\label{Emerging Tasks}

LLM-as-a-judge is also applied to tasks once exclusive to humans, particularly in context-specific areas. A prominent task is in social intelligence, where models are presented with complex social scenarios requiring the understanding of cultural values, ethical principles, and potential social impacts~\cite{xu2024academically, zhou2023sotopia}. Research has also extended to evaluating Large Multimodal Models (LMMs) and Large Vision-Language Models (LVLMs)~\cite{zhu2024adaptive}. For example,~\citet{xiong2024llava} use LMM-as-a-judge to provide transparent evaluations with rationales, while~\citet{chen2024automated} propose a benchmark for LVLMs in self-driving scenarios, showing that LLM-based evaluations align better with human preferences than LVLM-based ones. Recently, we have seen more customized utilization of LLM-as-a-judge to evaluate emerging tasks such as code understanding and generation~\cite{zhao2024codejudge,zhuo2024ice,tseng2024codev,wu2024can,he2025code,yureasoning,wang2025can,prasad2025learning,liu2025iterative,chi2025copilot}, legal knowledge~\cite{fei2023lawbench}, game development~\cite{isaza2024prompt}, nature science~\cite{bi2023oceangpt,chuang2025judging,kim2025towards}, manufacture engineering~\cite{liu2025llm}, healthcare conversations~\cite{wang2024healthq,zhang2024comprehensive,zhou2024llm}, debating judgment~\cite{liang2024debatrix}, RAG~\cite{Dhole2024ConQRetBF,saad2024ares,jin2024rag,liu2025judge,seo2025mt}, biomedical application~\cite{brake2024comparing,zheng2025hierarchical,zhang2024ace}, paper review~\cite{zhou2024llma,wang2024automated,zhu2025deepreview,kirtani2025revieweval}, novelty \& creativity evaluation~\cite{olson2024steering,feng2025grapheval,sawicki2025can}, and human-computer interaction~\cite{li2024iqa}.

\subsection{Alignment}
Alignment tuning is a vital technique to align LLMs with human preferences and values~\cite{weifinetuned, ouyang2022training, rafailov2024direct}. In this section, we discuss the use of larger LLMs as judges (Section~\ref{Larger Models as Judges}) and self-judging (Section~\ref{Self-Judging}) for alignment.

\subsubsection{Larger Models as Judges}
\label{Larger Models as Judges}

Recently, alignment tuning leverages feedback from larger LLMs to guide smaller models.~\citet{bai2022constitutional} first propose to train reward models with synthetic preferences from pre-trained LLMs. 
Following this, there are also some works explore online learning~\cite{guo2024direct} and direct preference optimization~\cite{lee2023rlaif} with larger models as judges.
To prevent reward hacking,~\citet{sun2024salmon} develop an instructable reward model enabling real-time human interventions for alignment. Moreover, multi-agent collaborations employ diverse workflows and LLM debates to improve judgments in alignment tuning~\cite{arif2024fellowship, Sengupta2024MAGVAM, li2024coevol}. For code alignment,~\citet{weyssow2024codeultrafeedback} create CodeUltraFeedback, a dataset using LLM judges to align smaller code models.~\citet{wang2024bpo} introduce BPO, employing GPT-4 as a judge to augment pairwise feedback.

\subsubsection{Self-Judging}
\label{Self-Judging}

Self-judging utilizes LLMs’ own preference signals for self-improvement.
Some focus on directly judging the preference ranking with the policy LLMs.
~\citet{yuan2024self,zhang2025process} first introduce self-rewarding, where LLMs judge their outputs to construct pairwise data. 
Following works adopt various methods to improve the judging capabilities, including meta-rewarding~\cite{wu2024meta}, Judge-Augmented Supervised Fine-Tuning (JSFT)~\cite{lee2024aligning} and self-evaluation~\cite{zhang2024self}.
To guarantee the quality of synthetic pairwise data,~\citet{pace2024west} introduce West-of-N approach while~\citet{tong2024optimizing} apply self-filtering to produce high-quality synthetic data pairs for reasoning tasks. 
To reduce computational overhead,~\citet{zhai2024online} propose ranked pairing for self-preferring models.~\citet{liu2024meta} introduce meta-ranking, enabling smaller LLMs to act as judges and combining this method with Kahneman-Tversky optimization for post-SFT alignment. Besides pairwise data,~\cite{liang2024sheep} and~\cite{yasunaga2024alma} leverage LLM-as-a-judge to filter synthetic instruction tuning data.
Other works adopt self-assessment and self-judgment in specific domains, such as robotics~\cite{zeng2024learning,yi2024protocollm} and multimodal~\cite{ahn2024srt}.

\subsection{Retrieval}

In traditional retrieval, LLM-as-a-judge ranks documents by relevance with minimal labeled data (Section~\ref{Traditional Retrieval}). LLM judges can also enhance the RAG system by dynamically integrating retrieved knowledge into the final response (Section~\ref{Retrieval-Augmented Generation (RAG)}). 

\subsubsection{Traditional Retrieval}
\label{Traditional Retrieval}

LLMs enhance document ranking by employing methods like permutation-based ranking~\cite{sun2023chatgpt}, fine-grained relevance labeling~\cite{zhuang2024beyond}, and listwise reranking without task-specific training~\cite{ma2023zero}. Moreover, Setwise~\cite{zhuang2024setwise} and Pairwise Ranking Prompting (PRP)~\cite{qin2024large} offer a cost-efficient alternative for complex tasks.~\citet{tang2024found} introduce a permutation self-consistency technique that averages across multiple orders to obtain order-independent rankings. Domain-specific knowledge retrieval with LLM-as-a-judge includes legal information, recommender systems and searching~\cite{ma2024leveraging, hou2024large, thomas2023large}.

\begin{table*}[h!]
\centering
\small
\begin{tabular}{p{4cm} p{10cm}}
\toprule[1.2pt]
\textbf{Benchmark} & \textbf{Definition} \\
\midrule
General Performance & Benchmarks that assess the general accuracy performance of LLM judges (e.g., MT-Bench) \\
\addlinespace[4pt]
Bias Quantification & Benchmarks focused on measuring and analyzing biases in LLM judgments (e.g., CALM) \\
\addlinespace[4pt]
Challenging Performance & Benchmarks that test LLM judges on difficult or adversarial tasks designed to probe the limits of their evaluation capabilities (e.g., Arena-Hard) \\
\addlinespace[4pt]
Domain-Specific Performance & Benchmarks that measure LLM judges’ effectiveness in specific domains, such as biomedical, legal, and coding evaluation (e.g., \citet{raju2024constructing}) \\
\bottomrule[1.2pt]
\end{tabular}
\caption{Categories of benchmarks for evaluating LLM judges.}
\label{tab:categories}
\end{table*}

\subsubsection{Retrieval-Augmented Generation (RAG)}
\label{Retrieval-Augmented Generation (RAG)}

\citet{li-qiu-2023-mot} propose the Memory-of-Thought (MoT) framework, where LLMs store and recall reasoning to enhance response relevance.~\citet{tang2024self} introduce Self-Retrieval, an architecture integrating retrieval into document generation, enabling end-to-end IR within a single LLM. Similarly,~\citet{asai2024selfrag} develop SELF-RAG, combining retrieval with self-reflection to enhance response quality. In the domain of Q\&A,~\citet{rackauckas2024evaluating} present an LLM-based evaluation framework using synthetic queries to judge RAG agent performance.~\citet{zhang2024large} study LLMs’ ability to assess relevance versus utility. In the biomedical area, several studies explore the usage of LLM-as-a-judge for active and dynamic retrival~\cite{wang2024biorag} or retrieved knowledge filtering~\cite{jeong2024improving,li2024dalk}.

\subsection{Reasoning}

Reasoning is a critical aspect of LLMs because it directly affects their ability to solve complex problems. Recently, many studies leverage LLM-as-a-judge in reasoning path selection (Section~\ref{Reasoning Path Selection}) and external source utilization (Section~\ref{Reasoning with External Tools}).

\subsubsection{Reasoning Path Selection}
\label{Reasoning Path Selection}

While many complex reasoning and cognition structures emerges for LLMs' reasoning~\cite{yao2023tree,hao2023reasoning}, one crucial challenge is how to select a reasonable and reliable reasoning path or trajectory for LLMs to reason.
To achieve this, LLM-as-a-judge has been introduced.
Some works adopt the reasoner LLMs to perform self-assessment, alternatively executing reasoning and judging steps to achieve the best result~\cite{lahoti2023improving,creswell2022selection,xie2024self,kawabata2024rationale} or perform sample-level selection among a group of candidates~\cite{musolesi2024creative}.
Additionally, there are also many work train LLM-based verifiers, leveraging the judge LLM as the process reward model (PRM) to evaluate each state~\cite{lightmanlet,lightman2023let,setlur2024rewarding,zhang2024generative,ye2025uncertainty}.
Besides, there are also studies train critique-based LLM judges~\cite{xu2024large,ankner2024critique,yu2024self,wang2024direct,lancriticeval,xie2024improving} which provide fine-grained verbal feedback to boost the reasoning process.

\subsubsection{Reasoning with External Source}
\label{Reasoning with External Tools}

Selecting an appropriate external source to use is essential in the success of agentic LLM systems~\cite{xi2023rise,wang2024survey}.
Auto-GPT~\cite{yang2023auto} is the first to benchmark LLMs' performance in real-world decision-making scenarios.
Following them, many other works adopt LLM-as-a-judge in various external tool selection applications, including autonomous driving~\cite{sha2023languagempc}, reasoning structure selection~\cite{zhou2024self} and multi-modal area~\cite{zhao2024diffagent}.
In addition to selecting among external tools or APIs, LLM-as-a-judge has also been widely adopted as a controller in multi-agent systems, to selectively activate agents for a given problem~\cite{ong2024routellm} or to assess and manage message flow among a group of agents~\cite{liang2023encouraging,li2024smoa}.

\subsection{Definition of each LLM-as-a-judge Benchmark Category}
We provide the definition of each LLM-as-a-judge benchmark in Table~\ref{tab:categories}.

\onecolumn
\section{Taxonomy}
\label{Taxonomy}
\tikzstyle{my-box}= [
    rectangle,
    draw=hidden-draw,
    rounded corners,
    text opacity=1,
    minimum height=1.5em,
    minimum width=5em,
    inner sep=2pt,
    align=center,
    fill opacity=.5,
]
\tikzstyle{leaf}=[my-box, minimum height=1.5em,
    fill=pink!60, text=black, align=left,font=\scriptsize,
    inner xsep=2pt,
    inner ysep
=4pt,
]
\begin{figure*}[h!]

    \centering
    \resizebox{\textwidth}{!}{
        \begin{forest}
            forked edges,
            for tree={
                grow=east,
                reversed=true,
                anchor=base west,
                parent anchor=east,
                child anchor=west,
                base=left,
                font=\scriptsize,
                rectangle,
                draw=hidden-draw,
                rounded corners,
                align=left,
                minimum width=4em,
                edge+={darkgray, line width=1pt},
                s sep=3pt,
                inner xsep=2pt,
                inner ysep=3pt,
                ver/.style={rotate=90, child anchor=north, parent anchor=south, anchor=center},
            },
            where level=1{text width=4.5em,font=\scriptsize,}{},
            where level=2{text width=5.4em,font=\scriptsize,}{},
            where level=3{text width=6.4em,font=\scriptsize,}{},
            where level=4{text width=6.4em,font=\scriptsize,}{},
            [
        LLM-as-a-judge, ver
        [
    Attributes\\ (\S \ref{Attributes})
    [
        Helpfulness\\ (\S \ref{Helpfulness})
        [
            Constitutional AI~\cite{bai2022constitutional}{, }RLAIF~\cite{lee2023rlaif}{, }MT-Bench~\cite{zheng2023judging}{, }Just-Eval~\cite{lin2023unlocking}{, }Starling \\~\cite{zhu2024starling}{, } AUTO-J~\cite{li2024generative}{, } OAIF~\cite{guo2024direct}{, }, leaf, text width=41em
        ]
    ]
    [
        Safety \& Security \\ (\S \ref{Harmlessness})
        [
            LLaMA Guard~\cite{inan2023llama}{, }TRUSTGPT~\cite{huang2023trustgpt}{, }Moral Choice~\cite{scherrer2024evaluating}{, }
            SORRY-Bench\\ ~\cite{xie2024sorry}{, }
            FLASK \cite{ye2024flask}{, }R-judge~\cite{yuan2024r}{, }Do-not-answer~\cite{wang2024not}{, }RAIN~\cite{lirain}, leaf, text width=41em
        ]
    ]
    [
        Reliability \\ (\S \ref{Reliability})
        [
            FactScore~\cite{min2023factscore}{, }HALU-J~\cite{wang2024halu}{, }HalluJudge~\cite{luo2024halludial}{, }HalluQA~\cite{cheng2023evaluating}{, }
            SaySelf\\~\cite{xu2024sayself}{, } ~\cite{wei2024long}{, }Self-Alignment for Factuality~\cite{zhang2024self}{, }FaithScore~\cite{jing2024faithscore}{, }FENCE\\ ~\cite{xie2024improving}, leaf, text width=41em
        ]
    ]
    [
        Relevance \\ (\S \ref{Relevance})
        [
            LLM-Eval~\cite{lin2023llm}{, }MoT~\cite{li-qiu-2023-mot}{, }~\cite{abbasiantaeb2024can}{, }
            DALK~\cite{li2024dalk}{, }MJ-Bench\\~\cite{chen2024mj}{, }\cite{thomas2024large}{, }\cite{ma2024leveraging}{, }LLMRank~\cite{hou2024large}{, }LLM Evaluation~\cite{chiang2023can}{, }\\ \cite{yang2024toward}{, }\cite{chen2024mllmasajudge}{, }LLM-SQL-Solver~\cite{zhao2023llm}{, }\cite{lin2025evaluating}, leaf, text width=41em
        ]
    ]
    [
        Logic \\ (\S \ref{Feasibility})
        [
            RAP~\cite{hao2023reasoning}{, }ToT~\cite{yao2023tree}{, }Auto-GPT~\cite{yang2023auto}{, }GoT~\cite{besta2024graph}{, }Diffagent~\cite{zhao2024diffagent}{, }\\ Routellm\cite{ong2024routellm}{, } MAD~\cite{liang2023encouraging}{, } SMoA~\cite{li2024smoa}, leaf, text width=41em
        ]
    ]
    [
        Overall Quality \\ (\S \ref{Overall Quality})
        [
           \cite{gao2023human}{, }Just-Eval~\cite{lin2023unlocking}{, }ICE~\cite{Jain2023MultiDimensionalEO}{, }LLM-Eval~\cite{Lin2023LLMEvalUM}{, }GEMBA\\~\cite{Kocmi2023LargeLM}{, }KIEVAL~\cite{yu2024kieval}{, }OAIF~\cite{guo2024direct}{, }Comp-Analysis~\cite{zhang2024comprehensive}{, }
           LostITS\\~\cite{Huang2024LostIT}, leaf, text width=41em
        ]
    ]
]
        [
    Methodology \\ (\S \ref{Methodology})
    [
        Tuning (\S \ref{Tuning})
        [
            Data Source\\ (\S \ref{Data Source})
            [
                Manually-labeled \\ (\S \ref{Data Source})
                [
                    AttrScore~\cite{yue2023automatic}{, }
                    PandaLM~\cite{wang2024pandalm}{, }
                    InstructScore\\~\cite{xu2023instructscore}{, }
                    SELF-JUDGE~\cite{lee2024aligning}{, }                  
                    X-Eval~\cite{liu2024x}{, }
                    \\CritiqueLLM~\cite{ke2024critiquellm}{, }
                    FLAMe~\cite{vu2024foundational}{, }, leaf, text width=25em
                ]
            ]
            [
                Synthetic Feedback \\ (\S \ref{synthetic})
                [
                    JudgeLM~\cite{zhu2023judgelm}{, }
                    AUTO-J~\cite{li2024generative}{, }
                    Meta-Rewarding\\~\cite{wu2024meta}{, }
                    Self-Taught~\cite{wang2024self}{, }
                    HALU-J \\~\cite{wang2024halu}{, }
                    OFFSETBIAS\cite{park2024offsetbias}{, }
                    SORRY-Bench\\~\cite{xie2024sorry}{, }
                    LLaVA-Critic~\cite{xiong2024llava}{, }                    
                    PROMETHEUS2\\~\cite{kim2024prometheus}{, }
                    InstructScore~\cite{xu2023instructscore}{, }, leaf, text width=25em
                ]
            ]
        ]
        [
            Tuning Techniques \\ (\S \ref{Tuning Techniques})
            [
                Supervised F-Tuning \\ (\S \ref{sft})
                [
                    PerSE~\cite{wang2023learning}{, }
                    INSTRUCTSCORE~\cite{xu2023instructscore}{, }
                    CRITIQUE-\\LLM\cite{ke2024critiquellm}{, }
                    PandaLM~\cite{wang2024pandalm}{, }X-Eval~\cite{liu2024x}{, }\\AUTO-J\cite{li2024generative}{, }JudgeLM~\cite{zhu2023judgelm}{, }SORRY-Bench\\~\cite{xie2024sorry}{, }AttrScore\cite{yue2023automatic}{, }FLAMe~\cite{vu2024foundational}{, }\\PROMETHEUS2~\cite{kim2024prometheus}{, }SELF-JUDGE~\cite{lee2024aligning}{, }Critique-\\LLM~\cite{ke2024critiquellm}{, }X-Eval~\cite{liu2024x}{, }, leaf, text width=25em
                ]
            ]
            [
                Reinforcement Learning \\ (\S \ref{pl})
                [
                    HALU-J~\cite{wang2024halu}{, }
                    OFFSETBIAS~\cite{park2024offsetbias}{, }
                    Themis\\~\cite{hu2024themis}{, }
                    Meta-Rewarding~\cite{wu2024meta}{, }
                    Self-Taught\\~\cite{wang2024self}{, }PORTIA~\cite{li2024split}, leaf, text width=25em
                ]
            ]
        ]
    ]
    [
        Prompting\\ (\S \ref{Prompting})
        [
            Swapping Operation \\ (\S \ref{swapping})
            [
                MT-Bench~\cite{zheng2023judging}{, }
                RLAIF~\cite{lee2023rlaif}{, }
                SALMON~\cite{sun2024salmon}{, }SELF-JUDGE\\~\cite{lee2024aligning}{, }Starling~\cite{zhu2024starling}, leaf, text width=33em
            ]
        ]
        [
            Rule Augmentation \\ (\S \ref{rule})
            [
                 Constitutional AI~\cite{bai2022constitutional}{, }
                 MoT~\cite{li-qiu-2023-mot}{, }
               \cite{lahoti2023improving}{, }
                RLAIF\\~\cite{lee2023rlaif}{, }
                LRQ-Fact~\cite{beigi2024lrq}{, }
                AUTO-J~\cite{li2024generative}{, }
               \cite{bai2024benchmarking}{, }       
                \\\cite{gao2023human}{, }
                Prometheus~\cite{kim2023prometheus}{, }
                KIEVAL\cite{yu2024kieval}{, }
                CEB~\cite{wang2024ceb}{, }\\
               \cite{murugadoss2024evaluating}{, }
               \cite{liu2024calibrating}{, } 
                OAIF~\cite{guo2024direct}{, }
                SALMON~\cite{sun2024salmon}{, }
                \\SELF-JUDGE~\cite{lee2024aligning}{, }                
                DALK~\cite{li2024dalk}{, }
               \cite{qian2024large}{, }
                RevisEval\\~\cite{zhang2024reviseval}{, }
                LLM-as-a-personalized-judge~\cite{dong2024can}{, }
                \cite{li2024rule}{, }\cite{li2024decompose}, leaf, text width=33em
            ]
        ]
        [
            Multi-Agent \\Collaboration (\S \ref{multi})
            [
                PRD~\cite{li2023prd}{, }
               \cite{zhang2023wider}{, }
               \cite{wu2023large}{, }
                MPA~\cite{zhu2024dynamic}{, }
                JudgeLM\\~\cite{zhu2023judgelm}{, }
                ChatEval\cite{chanchateval}{, } 
                CoEvol~\cite{li2024coevol}
                LRQ-Fact~\cite{beigi2024lrq}{, }
                \\Cascaded Selective Evaluation\cite{jung2024trust}{, }
                Fellowship~\cite{arif2024fellowship}{, }
                MATEval~\cite{li2024mateval}{, }\\
                \cite{Zhang2024BreakingER}, leaf, text width=33em
            ]
        ]
        [
            Demonstration \\ (\S \ref{demon})
            [
                ICE~\cite{jain2023multi}{, }
                Little Giants~\cite{kotonya2023little}{, }
                ALLURE~\cite{hasanbeig2023allure}{, }
                MSoR\\~\cite{song2024can}, leaf, text width=33em
            ]
        ]
        [
            Multi-Turn\\ Interaction (\S \ref{multiturn})
            [
                LLM-as-an-examine~\cite{bai2023benchmarking}{, }
                KIEVAL~\cite{yu2024kieval}{, }
                Auto-Arena~\cite{zhao2024auto}{, }\\
               ~\cite{moniri2024evaluating}, leaf, text width=33em
            ]
        ]
        [
            Comparison\\ Acceleration (\S \ref{compare})
            [
                \cite{liu2023statistical}{, }
                OSP~\cite{zhai2024online}{, }
                Starling~\cite{zhu2024starling}{, }
                SELF-JUDGE~\cite{lee2024aligning}, leaf, text width=33em
            ]
        ]
    ]
]
        [
    Application \\ (\S \ref{Application})
    [
        Evaluation\\ (\S \ref{Evaluation})
        [
           \cite{bi2023oceangpt}{, }\cite{fei2023lawbench}{, }\cite{zhou2023sotopia}{, }\cite{wang2023can}{, }\cite{nan2023evaluating}{, }\cite{zheng2023judging}{, }\cite{gao2023human}{, }\\\cite{wu2023large}{, }\cite{cheng2023evaluating}{, }\cite{Lin2023LLMEvalUM}{, }\cite{mondorf2024beyond}{, }\cite{badshah2024reference}{, }
           \\\cite{bai2024benchmarking}{, }\cite{kumar2024decoding}{, }\cite{wang2024halu}{, }\cite{li2024salad}{, }
           \cite{xie2024sorry}{, }\cite{chanchateval}{, }\\\cite{moniri2024evaluating}{, }
           \cite{xia2024evaluating}{, }\cite{fatemi2024test}{, }\cite{parmar2024logicbench} {, }\cite{xu2024academically}{, }
           \cite{xiong2024llava}{, }\\\cite{chen2024automated}{, }\cite{zhao2024codejudge}{, }\cite{isaza2024prompt}{, }\cite{wang2024healthq}{, }\cite{zengevaluating}{, }\cite{yu2024xfinder}{, }\\\cite{Dhole2024ConQRetBF}{, }\cite{Yang2024EvaluatingAA}{, }\cite{Xu2024BenchmarkingLJ}{, }\cite{Wu2024CanLL}, leaf, text width=41em
        ]
    ]
    [
        Alignment\\ (\S \ref{Alignment})
        [
           \cite{bai2022constitutional}{, }\cite{lee2023rlaif}{, }\cite{sun2024salmon}{, }\cite{guo2024direct}{, }\cite{arif2024fellowship}{, }\cite{li2024coevol}{, }\cite{yuan2024self}{, }\\\cite{wu2024meta}{, }\cite{pace2024west}{, }\cite{lee2024aligning}{, }\cite{tong2024optimizing}{, }\cite{zhai2024online}{, }\cite{liu2024meta}{, }\\\cite{liang2024sheep}{, }\cite{zhang2024self}{, }\cite{zeng2024learning}{, }\cite{ahn2024srt}{, }\cite{weyssow2024codeultrafeedback}{, }\cite{wang2024bpo}{, }\\\cite{yasunaga2024alma}{, }\cite{Sengupta2024MAGVAM}, leaf, text width=41em
        ]
    ]
    [
        Retrieval\\ (\S \ref{Retrieval})
        [
           \cite{sun2023chatgpt}{, }\cite{thomas2023large}{, }\cite{ma2023zero}{, }\cite{tang2024found}{, }\cite{qin2024large}{, }\cite{ma2024leveraging}{, }\cite{hou2024large}{, }\\\cite{li-qiu-2023-mot}{, }\cite{tang2024self}{, }\cite{asai2024selfrag}~\cite{zhuang2024beyond}{, }\cite{rackauckas2024evaluating}{, }\cite{zhang2024large}{, }\\\cite{wang2024biorag}{, }~\cite{li2024dalk}{, }\cite{jeong2024improving}{, }\cite{zhuang2024setwise}{, }\cite{chen2024llms}, leaf, text width=41em
        ]
    ]
    [
        Reasoning\\ (\S \label{Reasoning})
        [
            \cite{yao2022react}{, }\cite{creswell2022selection}{, }\cite{wei2022chain}{, }\cite{yao2023tree}{, }\cite{yang2023auto}{, }\cite{sha2023languagempc}{, }\\\cite{hao2023reasoning}{, }\cite{zhou2024self}{, }\cite{lahoti2023improving}{, }\cite{liang2023encouraging}{, }\cite{li2024smoa}{, }\cite{besta2024graph}{, }\\\cite{ong2024routellm}{, }\cite{zhao2024diffagent}{, }\cite{kawabata2024rationale}{, }\cite{xie2024improving}{, }\cite{lightman2023let}{, }\cite{lirain}{, }\\\cite{setlur2024rewarding}, leaf, text width=41em
        ]
    ]
]
    ]
        \end{forest}
    }
    \caption{Taxonomy of research in LLM-as-a-judge that consists of judging attribution, methodology and application.}
    \label{categorization_of_reasoning}
\end{figure*}

\clearpage
\section{Tuning Methods}
\label{Tuning Methods}
\begin{table*}[h!]
\centering
\small
\renewcommand{\arraystretch}{1.2}
\resizebox{0.95\linewidth}{!}{
\begin{tabular}{m{2.3cm} >{\centering\arraybackslash}m{1.5cm} >{\centering\arraybackslash}m{1.3cm} >{\centering\arraybackslash}m{2.6cm} >{\centering\arraybackslash}m{1.3cm} >{\centering\arraybackslash}m{1.5cm} >{\centering\arraybackslash}m{2.6cm} m{1.8cm}}
\toprule[1.2pt]

\multirow{2}{*}{\centering\textbf{Method}} & \multicolumn{4}{c}{\centering\textbf{Data}} & \multicolumn{2}{c}{\centering\textbf{Tuning Method}} & \multirow{2}{*}{\centering\textbf{Base LLM}} \\
\cmidrule(lr){2-5} \cmidrule(lr){6-7}
& \centering\textbf{Source} & \centering\textbf{Annotator} & \centering\textbf{Type} & \centering\textbf{Scale} & \centering\textbf{Technique} & \centering\textbf{Trick} & \textbf{} \\
\midrule

\rowcolor{gray!20} AttrScore \cite{yue2023automatic} & Manual & Human & QA, NLI, Fact-Checking, Summarization & 63.8K & SFT & - & Multiple LLMs \\
PandaLM \cite{wang2024pandalm} & Manual & Human & Instruction Following & 300K & SFT & - & Multiple LLMs \\
\rowcolor{gray!20} AUTO-J \cite{li2024generative} & Synthetic & GPT-4 & Real-world Scenarios & 4K & SFT & - & LLaMA-2 \\
JudgeLM \cite{zhu2023judgelm} & Synthetic & GPT-4 & Instruction Following & 100K & SFT & - & Vicuna \\
\rowcolor{gray!20} Self-Judge \cite{lee2024aligning} & Manual & Human & Preference Learning & 65/57K & SFT & JSFT & LLaMA-2 \\
X-EVAL \cite{liu2024x} & Manual & Human & Dialogue, Summarization, Data-to-Text & 55K & SFT & Two-Stage Instruction Tuning & Flan-T5 \\
\rowcolor{gray!20} FLAMe \cite{vu2024foundational} & Manual & Human & Various Tasks & 5M+ & SFT & Multi-task Training & PaLM-2 \\
InstructScore \cite{xu2023instructscore} & Manual\& Synthetic & Human\& GPT-4 & Various Tasks & 20K & SFT & Meta-Feedback & LLaMA \\
\rowcolor{gray!20} CritiqueLLM \cite{ke2024critiquellm} & Manual & Human & Instruction Following, real-world scenarios & 5K & SFT & Prompt Simplify, Swapping Augmentation & ChatGLM3 \\
Meta-Rewarding \cite{wu2024meta} & Synthetic & LLaMA-3 & Preference Learning & 20K & Preference Learning & Meta-Rewarding & LLaMA-3 \\
\rowcolor{gray!20} Self-Taught Evaluator \cite{wang2024self} & Synthetic & Mixtral & Various Tasks & 20K & Preference Learning & Self-Taught & LLaMA-3 \\
HALU-J \cite{wang2024halu} & Synthetic & GPT-4o & Fact Extraction & 2.6K & Preference Learning & DPO & Mistral \\
\rowcolor{gray!20} OffsetBias \cite{park2024offsetbias} & Synthetic & GPT-4, Claude3 & Preference Learning & 8.5K & SFT & Debiasing Augmentation & LLaMA-3 \\
SorryBench \cite{xie2024sorry} & Synthetic & GPT-4 & Safety & 2.7K & SFT & - & Multiple LLMs \\
\rowcolor{gray!20} LLaVA-Critic \cite{xiong2024llava} & Synthetic & GPT-4o & Preference Learning & 113K & Preference Learning & DPO & LLaVA-v.1.5 \\
PROME-THEUS2 \cite{kim2024prometheus} & Synthetic & GPT-4 & Preference Learning & 300K & SFT & Joint Training, Weight Merging & Mistral \\
\rowcolor{gray!20} Themis \cite{hu2024themis} & Manual \& Synthetic & Human \& GPT-4 & Various Tasks & 67K & Preference Learning & Multi-perspective Consistency Verification, Rating-oriented DPO & LLaMA-3 \\
\toprule[1.2pt]
\end{tabular}
}
\caption{Overview of tuning methods in LLM-as-a-judge.}
\label{tab:tuning}
\end{table*}

\section{Benchmark}
\label{Benchmark}

\begin{table*}[h!]
\small
\centering
\resizebox{0.95\linewidth}{!}{
\renewcommand{\arraystretch}{1.2}
\begin{tabular}{>{\centering\arraybackslash}m{2.0cm}
                >{\centering\arraybackslash}m{2.2cm}
                >{\centering\arraybackslash}m{1.0cm}
                >{\centering\arraybackslash}m{1.5cm}
                >{\centering\arraybackslash}m{3.2cm}
                >{\centering\arraybackslash}m{3.8cm}
                }
\toprule[1.2pt]
\textbf{Method} & \textbf{Data Type} & \textbf{Scale} & \textbf{Reference} & \textbf{Metrics} & \textbf{Purpose} \\
\midrule

MT-Bench \cite{zheng2023judging} & Multi-turn Conversation & 80 & Human Expert & Consistency, Bias, Error & General Performance, Position/Verbosity/Self-enhancement Bias \\
\rowcolor{gray!20} Chatbot Arena \cite{zheng2023judging} & Single-turn Conversation & 30K & User & Consistency, Bias, Error & General Performance, Position/Verbosity/Self-enhancement Bias \\
CodeJudge-Eval \cite{zhao2024codejudge} & Code & 457 & Execution System & Accuracy, F1 & General Performance \\
\rowcolor{gray!20} JudgeBench \cite{tan2024judgebench} & Various Tasks & 70K & Human & Cohen’s kappa, Correlation & General Performance \\
SOS-BENCH \cite{penfever2024sosbench} & Various Tasks & 152K & Human & Normalized Accuracy & General Performance \\
\rowcolor{gray!20} LLM-judge-eval \cite{wei2024systematic} & Summarization, Alignment & 1K & Human & Accuracy, Flipping Noise, Position Bias, Length Bias & General Performance \\
DHP \cite{wang2024dhp} & Various Tasks & 400 & Human & Discernment Score & General Performance \\

\rowcolor{gray!20} EvalBiasBench \cite{park2024offsetbias} & Alignment & 80 & Human & Accuracy & Various Bias \\


\citet{raju2024constructing} & Various Tasks & 1.5K & Human & Separability, Agreement, BrierScore & Domain-specific Performance \\


\rowcolor{gray!20} MLLM-as-a-judge \cite{chen2024mllmasajudge} & Various Tasks & 30K & Human & Human Agreement, Analysis Grading, Hallucination Detection & Multimodal \\


MM-EVAL \cite{son2024mmeval} & Various Tasks & 5K & Human & Accuracy & Multilingual \\
\rowcolor{gray!20} KUDGE \cite{son2024llm} & Question Answering & 3.3K & Human \& GPT-4o & Accuracy, Correlation & Non-English \& Challenging \\


\citet{murugadoss2024evaluating} & Various Tasks & - & Human & Correlation & Evaluation Instruction Following \\


\rowcolor{gray!20} \citet{thakur2024judging} & Question Answering & 400 & Human & Scott’s $\pi$, Percent Agreement & Vulnerability \\


Rewardbench \cite{lambert2024rewardbench} & Various Tasks & 20K & Human \& LLMs & Accuracy & General Performance \\
\rowcolor{gray!20} Arena-Hard Auto \cite{li2024crowdsourced} & Alignment & 500 & GPT-4-Turbo & Separability, Agreement & Challenging \\


R-Judge \cite{yuan2024r} & Multi-turn Interaction & 569 & Human & F1, Recall, Spec, Effect & Safety \\


\rowcolor{gray!20} \citet{shi2024optimization} & Alignment & 100K & Human & Repetition Stability, Position Consistency, Preference Fairness & Position Bias \\

CALM \cite{ye2024justice} & Various Tasks & 14K & Human & Robustness/Consistency Rate, 0riginal/ Hacked Accuracy & Bias Quantification \\

\rowcolor{gray!20} VL-RewardBench \cite{li2024vlrewardbench} & Various Tasks & 1.2K & Human \& LLMs & Overall Accuracy, Macro Average Accuracy & Multimodal \\

\toprule[1.2pt]
\end{tabular}
}
\caption{Overview of various benchmarks and datasets for LLM-as-a-judge.}
\label{tab:benchmark}
\end{table*}

\section{AI Assistants In Writing}
We acknowledge the use of ChatGPT-4o in paper polishing, but not in any direct paper writing or relevant work collections.

\end{document}